# Localized Observation Abstraction Using Piecewise Linear Spatial Decay for Reinforcement Learning in Combat Simulations

| | |
|---|---|
| **Scotty Black** | **Christian Darken** |
| Naval Postgraduate School | Naval Postgraduate School |
| Monterey, California | Monterey, California |
| scotty.black@nps.edu | cjdarken@nps.edu |

## ABSTRACT

In the domain of combat simulations, the training and deployment of deep reinforcement learning (RL) agents still face substantial challenges due to the dynamic and intricate nature of such environments. Unfortunately, as the complexity of the scenarios and available information increases, the training time required to achieve a certain threshold of performance does not just increase, but often does so exponentially. This relationship underscores the profound impact of complexity in training RL agents. This paper introduces a novel approach that addresses this limitation in training artificial intelligence (AI) agents using RL. Traditional RL methods have been shown to struggle in these high-dimensional, dynamic environments due to real-world computational constraints and the known sample inefficiency challenges of RL. To overcome these limitations, we propose a method of localized observation abstraction using piecewise linear spatial decay. This technique simplifies the state space, reducing computational demands while still preserving essential information, thereby enhancing AI training efficiency in dynamic environments where spatial relationships are often critical. Our analysis reveals that this localized observation approach consistently outperforms the more traditional global observation approach across increasing scenario complexity levels. This paper advances the research on observation abstractions for RL, illustrating how localized observation with piecewise linear spatial decay can provide an effective solution to large state representation challenges in dynamic environments.

## ABOUT THE AUTHORS

**Scotty Black** is a Lieutenant Colonel in the U.S. Marine Corps assigned to the Naval Postgraduate School as part of the Marine Corps Technical PhD Program. His primary specialty is as an F/A-18 Weapons Systems Officer with over 18 years of Marine Corps experience. LtCol Black is currently a PhD Candidate conducting research leveraging hierarchical reinforcement learning to scale artificial intelligence to deal with the often large and complex state spaces inherent in combat modeling and simulation for wargaming. LtCol Black's experience includes nearly 2,000 flight hours in the F/A-18, a graduate of the Weapons and Tactics Instructor (WTI) Course, multiple combat deployments, leading science and technology initiative for Marine Corps training and education as a Modeling and Simulation Officer, a DARPA Service Chiefs Fellowship, and research fellowships at the Naval Information Warfare Center Pacific and the former Space and Naval Warfare Systems Center Pacific.

**Christian Darken** is an Associate Professor in the Department of Computer Science at the Naval Postgraduate School, where he is also a member of the MOVES (Modeling, Virtual Environments and Simulation) faculty. He has more than 35 years of machine learning research experience and has been conducting teaching and research on human behavior models for simulations for over twenty years. His background includes technical program management for Siemens Corporation, and serving as program and general chair for the AAAI-sponsored AIIDE conference.

# Localized Observation Abstractions Using Piecewise Linear Spatial Decay for Deep Reinforcement Learning in Combat Simulations

| | |
|---|---|
| Scotty Black | Christian Darken |
| Naval Postgraduate School | Naval Postgraduate School |
| Monterey, California | Monterey, California |
| scotty.black@nps.edu | cjdarken@nps.edu |

## INTRODUCTION

In the domain of combat simulations, the training and deployment of deep reinforcement learning (RL) agents still face substantial hurdles due to the dynamic and complex nature of such environments, which often results in an exponential increase in training time required to achieve a certain threshold of performance. Traditional RL approaches often struggle to learn in these environments due to the large state spaces necessary to properly represent the number of entities, detailed terrains, and variable initial starting conditions characteristic of wargaming. This complexity, compounded by RL's sample inefficiency, vastly increases the computational resources and time-to-train needed to achieve satisfactory agent performance outcomes—rendering the process impractically costly and time-consuming.

This paper presents a novel approach to overcome these challenges by abstracting the agent's observation space while preserving sufficient detail of the relevant portions of the environment. By abstracting the state space into a more compact and computationally manageable observation while still maintaining critical spatial information, we aim to enhance training efficiency while significantly reducing the computational load needed. Specifically, our investigation delves into optimizing training efficacy against the backdrop of limited computational budgets—a common constraint in applying RL to combat simulations.

In this study, we develop, implement, and test a localized observation abstraction approach using piecewise linear spatial decay. Through our analysis, we demonstrate that a localized observation strategy consistently outperforms a global observation method across increasing levels of complexity. This finding emphasizes the superiority of this approach when training agents in complex scenarios where spatial relationships are essential—offering a way to help RL scale to still produce acceptable levels of performance in larger, more dynamic environments than have previously been possible in the domain of combat simulations.

## BACKGROUND

**Reinforcement Learning**

Reinforcement learning is a subset of machine learning that involves an agent learning to make decisions through direct interaction with its environment. In this process, the agent executes actions and receives feedback, either as rewards or penalties. Its objective is to optimize the cumulative reward over time. Through this method of trial-and-error search, the agent gradually learns the most effective actions based on the current state of the environment. This accumulated knowledge forms the agent's policy, which is essentially the set of strategies it uses for decision-making.

More formally put, a reinforcement learning problem, shown in Figure 1, typically consists of a decision-maker, referred to as an *agent*, and an *environment*, represented by *states* $s \in S$. The *agent* takes *actions* $a_t$ as a function of the current *state* $s_t$ such that $a_t = A(s_t)$. After choosing an *action* at time $t(a_t)$, the agent receives a *reward* $r_{t+1}$ and finds itself in a new *state* $s_{t+1}$. The *action* $a_t$ comes from a strategy called a *policy* $\pi$ that maps *states* $s \in S$ to a probability of selecting each possible *action* $\pi(s, a)$. As the *agent* interacts with the *environment*, it learns the optimal *policy* that maximizes its *reward* in the long run.

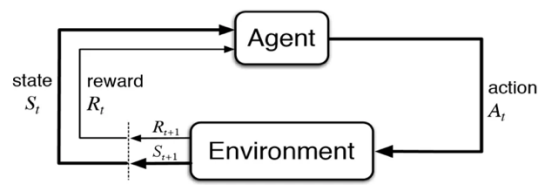

**Figure 1. The Reinforcement Learning Problem**

Although in this study we use RL to train intelligent agents in a combat simulation environment, this paper focuses on abstracting the state $s_t$ in a way that makes the RL problem more tractable in complex environments. Research we have leveraged in our approach to RL specifically include those exploring Atari 2600 games (Mnih et al., 2015; Van Hasselt et al., 2016), Go (Holcomb et al., 2018), Chess (Silver et al., 2017), Shogi (Silver et al., 2017), Dota 2 (Berner et al., 2019), StarCraft II (Vinyals et al., 2019), and Atlatl (Allen, 2022; Boron, 2020; Cannon & Goericke, 2020). Nevertheless, despite RL having achieved human, expert, or even superhuman-level play in some of these games, to date, AI agents have not been shown to significantly outperform humans or scripted (rule-based) agents in the complex domain of wargaming. In fact, research applying RL to combat simulations has shown that despite good outcomes in small scenarios, scaling to larger scenarios has typically resulted in poor performance (Boron, 2020; Cannon & Goericke, 2020; Rood, 2022). We posit that this is in large part due to the exponential growth of state space complexity (Bellman, 1954) and well-documented sample inefficiency problem in RL requiring extensive interaction with the environment (Mnih et al., 2015), especially as its observation space grows in complexity.

**State Abstraction**

The notion of abstraction for AI is not new and has been used since the beginning of AI and logic, dating back to the work by Whitt in approximating dynamic programs (Whitt, 1978; Abel, 2019). Giunchiglia and Walsh (1992) informally defined *abstraction* as "the process of mapping a representation of a problem onto a new representation." Abstraction allows people to consider what is relevant and to forget or disregard what is irrelevant based on the specific task at hand (Giunchiglia & Walsh, 1992).

Within the context of AI and RL, abstraction plays a critical role in simplifying complex decision-making processes. Given limited computational power and a complex enough environment, agents in a simulation cannot model everything in their environment and still learn appropriate or optimal behaviors within a reasonable time. As the complexity of the environment increases—and assuming that a minimum level of performance is desired—agents may have to discard some information and focus only on relevant information to solve a specific problem. This form of abstraction allows for a more manageable representation of intricate environments, enhancing the learning efficacy of AI agents (Ho et al., 2019). Moreover, this approach not only has the potential to reduce computational demands but may also improve the adaptability and performance of AI-trained agents in scenarios that may be significantly different from the scenarios for which the agents were trained (Abel, 2019).

As Shanahan and Mitchell (2022) explore in depth in their research, for abstraction to be most useful, "the domain of a concept's application must be larger than the domain of its acquisition." We contend that abstraction is critical to transferring concepts learned from one setting to another that differs from which it acquired said concept. Nevertheless, because abstractions inherently discard information—potentially compromising the effectiveness of the decisions made based on these abstractions—we must understand and balance the trade-off between making learning easier (or tractable) and preserving enough information to allow for optimal policy discovery (Abel, 2020). The more we abstract the state space, the more information is lost and the harder it will be to guarantee an optimal or near-optimal solution (Li et al., 2016). Nevertheless, researchers agree that a tradeoff exists in that, although coarser abstractions may result in sub-optimal actions, they allow for better planning and value iteration (Li et al., 2016).

While the concept of abstraction as applied to RL has slowly evolved, Abel (2019) formalizes this notion and comprehensively investigates the role of abstraction in RL in detail, particularly focusing on state abstraction. For this paper, we use Abel's (2020) definition of *State Abstraction* as a function $\phi : S \rightarrow S_\phi$, which maps each true environmental state $s \in S$ into an abstract state $s_\phi \in S_\phi$. In other words, an abstracted state serves as the agent's interpretation of the current environment, which will discard or simplify some information.

**RELATED WORKS**

The field of abstraction in reinforcement learning (RL) has seen a variety of approaches, each addressing different aspects of complexity in the decision-making processes. Understanding these works contextualizes our research and highlights the gaps our study aims to fill.

Sutton et al. (1999) pioneered the concept of temporal abstraction in RL by extending Markov decision processes (MDPs) (Puterman, 1994) and proposing semi-Markov decision processes (SMDPs). This foundational work emphasized understanding temporal factors in decision-making, focusing on the abstraction of actions rather than states. While crucial in developing the RL framework, it differs from our approach, which instead concentrates on state abstraction in the spatial context.

Further exploring abstraction in games with large state spaces, Sandholm (2015) introduced sophisticated methods for game-theoretic abstraction in large incomplete-information games. His work involved creating simpler models of games that maintained strategic similarity. This methodology is instrumental in game theory but diverges from our focus on state abstraction tailored to specific spatial dynamics.

Andersen et al.'s (2018) study on variational autoencoders (VAEs) captures elements of our approach in simplifying complex state spaces. Their emphasis, however, is on probabilistic latent space encoding as opposed to our deterministic spatial representation. While VAEs provide valuable insights into data encoding, our method focuses on explainable spatial relationships vital to decision-making in combat-like scenarios.

Ho et al. (2019) highlighted the critical role of abstraction in AI and RL, especially in managing complex environments. Their insights into state and temporal abstractions for efficient decision-making align closely with our work. Ho et al. (2019) demonstrated how abstraction simplifies computations and facilitates efficient trade-offs in learning—informing our approach in combat simulations in support of wargaming.

In a more focused application, Allen et al. (2021) utilized Markov processes for state space compression in RL. Specifically, their method grouped similar states based on transition patterns. While this offers a useful form of abstraction, it contrasts significantly with our approach where the spatial component, rather than the transition patterns, is central in informing optimal behaviors.

Lastly, Jergeus et al. (2022) took a unique approach by proposing linguistic abstractions in RL using a neuro-symbolic framework. While a completely different form of abstraction, their focus on abstracting linguistic communication among agents illuminates the versatility of abstraction techniques.

Each of these works illustrates the broader application of abstraction in RL. Collectively, they demonstrate the diverse methods of tackling complexity in the decision-making process. Our research builds upon these foundations, explicitly addressing the underexplored area of spatial state abstraction in the complex and intricate domain of combat simulations. As discussed by Abel (2020) in his Ph.D. Dissertation, *A Theory of State Abstraction for Reinforcement Learning*, RL agents currently face significant challenges in generalizing experiences, exploring environments, and learning from delayed and sparse feedback, all within limited computational constraints. Abel (2020) highlights the necessity of abstraction in these processes, focusing on state abstraction, to improve sample efficiency in RL. Furthermore, he outlines three desiderata for useful state abstraction—preserving near-optimal behavior, being learnable and computable efficiently, and reducing the time or data needed for effective decision-making—all of which we also seek to achieve in this study.

**METHODOLOGY**

To compare the tradeoffs between the traditional global observation approach and our approach using localized observation abstraction with piecewise linear spatial decay, we use the Atlatl simulation environment and employ the following methodology to investigate the tradeoff in agent performance vs. scenario complexity.

**Atlatl Simulation Environment**

We use the Atlatl Combat Simulation environment (Darken, 2022) to develop, implement, and experiment with our research approach. Atlatl is a simple but effective combat model developed at the Naval Postgraduate School (NPS). It includes an underlying combat model that is purposefully simplistic, as well as the surrounding Gymnasium (Farama Foundation, 2023) infrastructure that supports rapid AI experimentation. The environment also contains hooks that enable interfacing with standard RL codebase and algorithms, such as Stable-Baselines 3 (SB3). This type of basic environment allows researchers to develop, apply, and evaluate cutting-edge AI to operational and tactical problems more efficiently and effectively than using operational or high-fidelity simulation systems.

Through a web browser interface, Atlatl allows a human player to play against the AI; however, the simulation can also run headless with an AI playing against another AI. Additionally, a browser-based replay capability allows for replays of AI versus AI engagement. An example of a simple scenario on an Atlatl game board is shown in Figure 2. Units are represented visually by their respective military operational terms and graphics, and terrain is represented visually in colors (e.g., water is blue, rough terrain is brown, urban is gray).

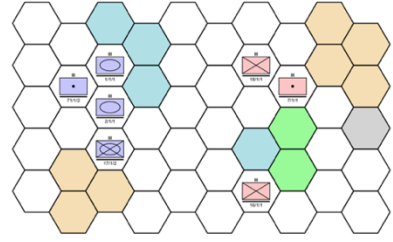

**Figure 2. Atlatl Gameboard Example**

The scoring system within Atlatl is fully customizable. Currently, performance within Atlatl is scored based on kills, losses, and holding urban areas, though these metrics can easily be changed as desired. Game scores are given from the perspective of the blue player. For this experiment, scoring is primarily computed based on two factors: combat effectiveness and control of urban areas (used to represent cities). We use the following scoring function:

$$S_{total} = S_{blue\_city} + S_{blue\_combat} - (S_{red\_city} + S_{red\_combat})$$

Where $S_{blue\_city}$ is the score per city owned by the blue faction; $S_{blue\_combat}$ is the score for each red agent damaged in combat by the blue faction; $S_{red\_city}$ is the score per city owned by the red faction, and $S_{red\_combat}$ is the score for each blue agent damaged in combat by the red faction.

As shown, control of urban hexagons plays a significant role in calculating the player's score. At the start of each scenario, unoccupied urban hexagons are not controlled by any faction. Control shifts only when an entity occupies the urban hexagon, with the controlling faction awarded a score of 24 points per phase controlled. Of note, once an urban hexagon is occupied, it remains under that faction's control even if the entity vacates the hexagon, up until an entity of the opposing faction occupies the same urban hexagon.

Regarding combat, each entity begins with an initial 100 strength points. Each damage point inflicted on a red entity translates into a positive point for the blue faction, while each damage point inflicted on a blue entity translates into a negative point for the blue faction. If an entity's strength drops below 50 points, it is removed from the game (i.e., deemed ineffective) and the remaining strength points are awarded to the opposing faction.

**Global Observation**

The global observation in Atlatl consists of an $18 \times n \times m$ tensor, where *n* and *m* are the height and width of the gameboard. For this study, we use square gameboards (e.g., a $5 \times 5$ scenario consists of an observation space of $18 \times 5 \times 5$). Each channel of the tensor represents one specific type of information to be captured, as shown in Figure 3. Specifically, channel 0 is a binary matrix depicting where the blue unit to be moved (or on-turn) is located; channel 1 is binary matrix depicting all blue units that still have the ability to move during the current phase; channel 2 is a binary matrix depicting all legal moves available for the unit on-move; channels 3 and 4 are matrices that depict the health level (scaled from 0 to 1.0) of each respective unit on the gameboard based on factions; channels 5 through 8 are binary matrices depicting unit types; channels 9 through 13 are binary matrices representing terrain; channels 14 and 15 are binary matrices depicting the city owner (i.e., which faction was the last to pass through an urban hexagon); channel 16 is a matrix filled with a phase indicator value representing the current phase of the game; and channel 17 is a matrix filled with the normalized game score. While we recognize that these last two features can be represented more compactly as vectors or scalars rather than matrices, we maintain the matrix construct for simplicity.

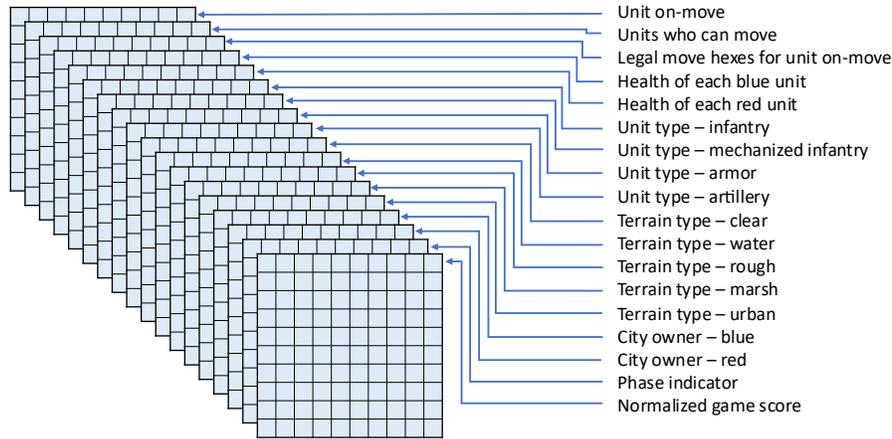

**Figure 3. Global Observation Image Channels**

**Localized Observation Abstraction Using Piecewise Linear Spatial Decay**

Our localized observation space takes in the game's global observation space described above and performs additional processing to compress the information into an $18 \times 7 \times 7$ observation, regardless of actual gameboard size. Even gameboards smaller than $7 \times 7$ are represented as a $7 \times 7$ with the area outside of the gameboard simply represented with zeros. To construct the localized $7 \times 7$ matrix, we first center the global matrix on the agent on-move. We then divide the entire area into 24 equal segments (due to the outer perimeter of this matrix consisting of 24 total grids) of 15° each. Finally, we multiply each entry by a weight $w$ as a function of Euclidean distance $d$, determined by the following equation, and visually depicted in Figure 4:

$$w(d) = \begin{cases} 1 & \text{for } d \leq 3 \\ 1 - 0.9 * \left(\dfrac{d-3}{7-3}\right) & \text{for } 3 < d < 7 \\ 0.1 - 0.9 * \left(\dfrac{d-7}{100-7}\right) & \text{for } 7 \leq d < 100 \\ 0.01 & \text{for } d \geq 100 \end{cases}$$

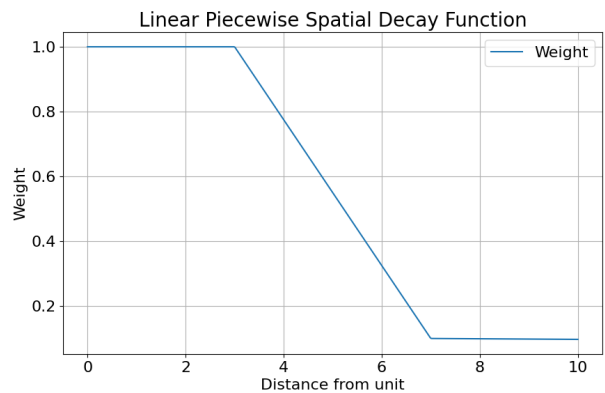

**Figure 4. Visual Depiction of Piecewise Linear Decay Function**

A conceptual illustration of this method is shown in Figure 5. Moving from left to right, the first image in Figure 5 is an example $10 \times 10$ gameboard with a single urban hexagon, 3 blue units, and 4 red units. The second image shows the inner $5 \times 5$ grid overlay in blue. Everything within this $5 \times 5$ will remain to scale due to its multiplication by a weight of 1. The third image depicts the area in which each element is first multiplied by a linearly decaying weight, then summed with the other values found within the respective 15° radial sector, and finally inserted into the respective location in the outer perimeter layer of the 7 × 7 localized grid. For each channel in the observation, this results in a 7 × 7 grid where the inner 5 × 5 is the 2 layers of adjacent hexagons surrounding the unit on-move to scale, and the outer perimeter of the 7 × 7 grid represents all of the information from the rest of the gameboard compressed by radial sector into a single value. To prevent the convolutional kernels from being distorted by disproportionately high values on the perimeter, we allow a maximum value of 1.0. This measure ensures that the data across both inner and perimeter hexagons is kept within a uniform scale, allowing the convolutional network to accurately interpret spatial relationships and maintain consistent performance across the gameboard.

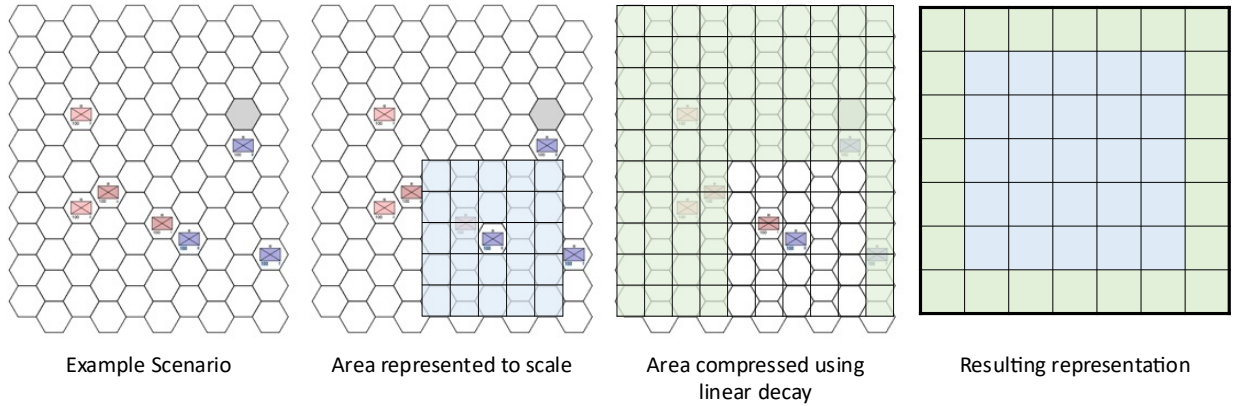

Example Scenario     Area represented to scale     Area compressed using linear decay     Resulting representation

**Figure 5. Graphical Depiction of Localized Observation**

It must be noted that fidelity of information is lost in this process. For example, a unit at 100 strength points 10 hexagons away would be represented with a value of 0.7, whereas a unit at 50 health points that is 4 hexagons away would be represented with a value of 0.775. Furthermore, because we are also summing up by radial sector, the resulting value can be misleading or ambiguous as it conflates separate data points into a single value—potentially confounding the interpretation of individual unit strengths. Nevertheless, we postulate that this approach still provides the agent with sufficient information to make optimal or near-optimal near-term decisions, while still maintaining awareness of the entire gameboard.

Within the context of Abel's (2020) definition of *State Abstraction* $\phi : S \rightarrow S_\phi$, $\phi$ is our function that takes the global state $s \in S$, applies our localized piecewise linear spatial decay method described above, and converts it into an abstracted state $s_\phi \in S_\phi$.

**Experiment**

Using the Atlatl simulation environment, we design and conduct the following experiment.

*Gymnasium Environment*
For our RL training, we use a custom Gymnasium (Farama Foundation, 2023) environment configurable for different roles ("blue" or "red"), AI types, and scenarios. The action space of our RL agent is defined as 7 discrete actions, one for each adjacent hexagon, plus the option to "pass" (i.e., take no action). Legal moves are defined as either moving to an unoccupied adjacent hexagon or engaging in combat by selecting a hexagon occupied by a unit of the opposing faction.

*Neural Network Architecture*
We use a residual convolutional neural network (CNN) specifically designed to process a hexagonally structured input observation space of any size which, for this study, is $18 \times n \times n$ where $n$ is the size of one side of the gameboard. The architecture uses convolutional layers to transform the input observation tensor into **64** output channels. This is followed by **7** additional layers of **64** channels each. Each layer features HexagDLy hexagonal convolutions (Steppa & Holsch, 2019) with a kernel size of $1 \times 1$ and a stride of $1$. Additionally, in each layer, we include a Rectified Linear Unit (ReLU) activation function and a residual connection. After **7** layers, the resulting multi-dimensional tensor is then flattened into a one-dimensional tensor and is passed through a final linear layer. This layer maps the flattened tensor to a **512**-dimensional feature vector, which is then passed through a final **ReLU** activation function.

*Reinforcement Learning Algorithms*
We employ the Deep Q-Network (DQN) algorithm (Raffin, 2018). The hyperparameters used were optimized through extensive hyperparameter tuning in similar scenarios, though not specific to this experiment. The final configuration included a learning rate of $\mathbf{0.0002}$, a buffer size of $\mathbf{1,000,000}$, learning starting at $\mathbf{10,000}$ steps, a batch size of **64**, and a discount factor ($\boldsymbol{\gamma}$) of $\mathbf{0.93}$. The target network update interval was set to $\mathbf{1,000}$ steps. For exploration, we employed an initial epsilon ($\boldsymbol{\varepsilon_i}$) of $\mathbf{1.0}$, decaying linearly to a final epsilon ($\boldsymbol{\varepsilon_f}$) of $\mathbf{0.01}$, with an exploration fraction of $\mathbf{1.0}$. The training frequency was set to every **4** steps, with a gradient step of **1** per training update.

*Scenarios*
We use randomly generated scenarios consisting of square hexagonal gameboards with one city and no other terrain. We use scenario gameboard sizes from $3 \times 3$ up to $12 \times 12$ in increments of 1, with each representing an increase in complexity level, where complexity level 3 is represented by a $3 \times 3$, complexity level 4 by a $4 \times 4$, and so on. Examples are shown in Figure 6. Each game begins with a random number of entities per faction with a minimum and maximum number computed as a factor of the length of the gameboard, where $num\_units_{max} = gameboard\_length$ and $num\_units_{min} = \text{round}(\frac{gameboard\_length}{2})$. For example, for a $5 \times 5$ gameboard, the scenario would start with a random number of units per faction between 3 and 5; whereas for a $10 \times 10$ scenario, the random number of starting units per faction would be between 5 and 10. Each scenario also includes 1 urban hexagon randomly placed according to force ratio. If one faction has a smaller force ratio (i.e., less units as compared to the opposing faction), the city is placed on their side of the gameboard. If the force ratios are equal (i.e., both factions have an equal number of units), the city is placed in a neutral location along the middle axis of the board. We set the number of phases in the game as $phases = 4 * gameboard\_length$, where each phase is one entire turn for one faction (i.e., one faction is allowed to make one legal move for each of its available entities). Setting the number of phases to this value provides enough turns for a unit to go from one end of the gameboard to the opposite end and return, likely giving them enough turns to execute complex maneuvering if warranted.

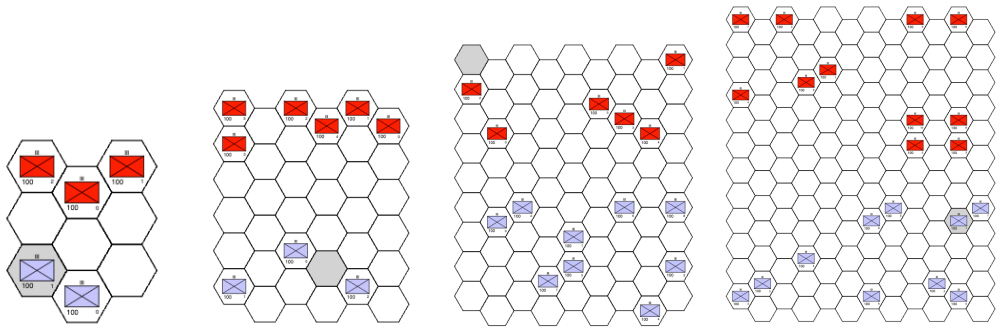

**Figure 6. Example Scenarios (From Left to Right, Complexity Levels 3, 6, 9, 12)**

*Training*
We train each model for 10 million steps against a baseline rule-based adversary model we call *Pass-Agg*. This name is derived from the terms "passive" and "aggressive." The agent first assesses its posture as "attack" or "defend" based on the relative strength of its faction as compared to its opponent. The agent prioritizes engaging any enemy units within its attack range of 1 hexagon (i.e., hexagons adjacent to its own position). If multiple targets exist within range, the agent uses a uniform distribution to select its target. In the absence of attack opportunities, the agent assesses which hexagon to move to based on proximity to enemy units and urban hexagons. The agent seeks to position itself advantageously while maintaining a balance between offensive actions and strategic repositioning. This decision is based on a hexagon scoring system that evaluates the advantage of moving towards urban hexagons or attacking nearby enemies. If neither attacking nor moving is advantageous, the agent may choose to "pass", effectively maintaining its current position. While a simple model, *Pass-Agg* has proven to be an effective agent that regularly achieves near-optimal scores and displays credible moves that would be expected of these combat units.

To learn effective behaviors, we design a reward system that balances defeating the opposing faction and occupying urban hexagons with preserving its own force. Our rewards are computed at each time step using the following equation:

$$R_{engineered} = \max(R_{raw}, 0)\frac{S_c}{S_o} + B_t I_t$$

Where $R_{raw}$ is the difference in game score between the current time step and the previous time step; $S_c$ is the current total friendly strength; $S_o$ is the original total friendly strength; $B_t$ is a terminal bonus reward of 25 points that our research shows discourages units from moving into the adversary units' attack range during the last turn of the game; and $I_t$ is a terminal game state indicator that takes on a value of 1 if the game is terminal or 0 if the game is not terminal.

*Evaluation*

We evaluate each of our trained models against the *Pass-Agg* model. We run 100,000 games where each game begins with a randomly generated scenario using the scenario parameters specified above. In addition to training with the *Pass-Agg* behavior model as the adversary, we also evaluate *Pass-Agg* vs. *Pass-Agg* as our rule-based model baseline. While we anticipate that Pass-Agg will outperform the RL-trained models as complexity levels increase, we aim to assess the extent of improvement an RL-trained agent offers over a rule-based agent while also seeking to determine when this relationship reverses. The performance of the *Pass-Agg* model serves as our benchmark to determine the point at which an RL-trained model ceases to surpass the effectiveness of a rule-based approach. Lastly, we also evaluate a random-actions model to determine when our RL-trained models do no better than, or converge to, a random actor.

**RESULTS AND DISCUSSION**

With each trained model, we run an evaluation consisting of 100,000 randomly generated games for each behavior model at each complexity level against our baseline *Pass-Agg* adversary behavior model. The means of the scores are presented in Table 1. For conciseness, in the following sections, we use the term *Local* to refer to our RL-trained model utilizing the localized observation abstraction using piecewise linear spatial decay; the term *Global* to refer to the RL-trained model using a global observation; the term *Rule-Based* to refer to the scripted *Pass-Agg* model; and the term *Random* to refer to the random-actions model. Overall, we see in Table 1 that *Local* outperforms *Global* across all complexity levels by a large margin. Furthermore, we also see the *Local* outperforms *Rule-Based* in complexity levels 3 through 5 by a large margin and, as expected, begins to fall off as complexity increases.

**Table 1. Mean ($\bar{x}$) Scores Across 100,000 Games for Each Model at Each Level of Complexity**

| | | Complexity | | | | | | | | | |
|---|---|---|---|---|---|---|---|---|---|---|---|
| | | 3 | 4 | 5 | 6 | 7 | 8 | 9 | 10 | 11 | 12 |
| Model | Local | 181.4 | 227.7 | 225.3 | 88.4 | -154.0 | -394.6 | -491.4 | -620.2 | -860.2 | -908.6 |
| | Global | 31.5 | -203.9 | -488.6 | -653.8 | -808.8 | -885.5 | -1039.0 | -1116.0 | -1229.9 | -1307.1 |
| | Rule-Based | 50.0 | 97.4 | 124.3 | 124.5 | 133.2 | 130.6 | 133.1 | 134.9 | 141.9 | 128.7 |
| | Random | -339.5 | -465.7 | -628.8 | -724.2 | -865.4 | -948.0 | -1078.6 | -1158.4 | -1284.7 | -1364.6 |

**Table 2. Raw Standard Error of the Mean (*SEM*) of Each Model's Scores Across 100,000 Games**

| | | Complexity | | | | | | | | | |
|---|---|---|---|---|---|---|---|---|---|---|---|
| | | 3 | 4 | 5 | 6 | 7 | 8 | 9 | 10 | 11 | 12 |
| Model | Local | 0.9 | 1.2 | 1.7 | 2.0 | 2.2 | 2.1 | 2.2 | 1.8 | 1.3 | 1.3 |
| | Global | 1.0 | 1.0 | 0.8 | 0.8 | 0.7 | 0.8 | 0.9 | 0.9 | 1.0 | 1.0 |
| | Rule-Based | 1.0 | 1.3 | 1.7 | 1.9 | 2.3 | 2.6 | 3.0 | 3.3 | 3.7 | 3.9 |
| | Random | 0.5 | 0.6 | 0.6 | 0.6 | 0.7 | 0.7 | 0.7 | 0.8 | 0.8 | 0.8 |

To verify that these differences in mean scores between our models are statistically significant, we set $\alpha = 0.05$ and run the Tukey-Kramer Honest Significant Difference (HSD) test. This test conducts pairwise comparisons between all possible pairs and details which specific groups' means are significantly different from each of the other groups. We find statistical significance between mean scores for each model across all complexity levels used in our experiment, each producing a p-value of $< .0001$. Additionally, we show the Standard Errors of the Mean (SEM) for each model based on complexity levels in Table 2.

Figure 7 depicts the line plot of *Mean Scores* by *Complexity* for each model evaluated. Figure 8 shows the same data normalized to the random-actions behavior model (*Random*) using the standard normalization formula $x_{\text{Norm}} = \frac{x - \bar{x}_{\text{Random}}}{\sigma_{\text{Random}}}$. We normalize to *Random* for visualization purposes as we contend that an untrained neural network should produce random actions and can, therefore, be representative of the reasonably worst-case score. However, we acknowledge that even with training, it is still possible for a neural network to do worse than a random actor. Nevertheless, we use *Random* as our baseline for our zero line for graphical purposes to better observe how and when *Global* and *Local* converge to *Random* (i.e., when our models perform no better than a random actor).

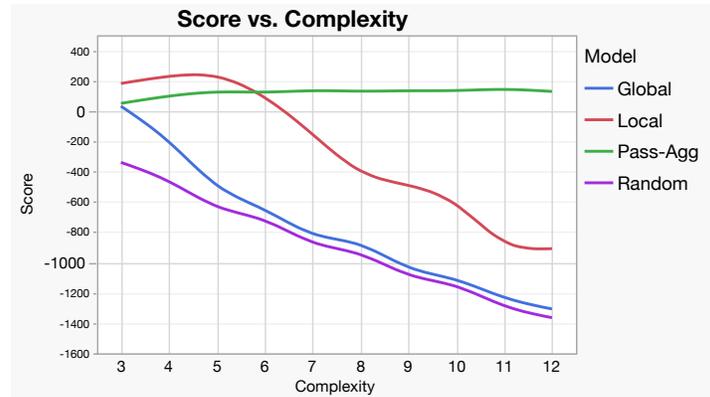

**Figure 7. Mean Score Vs. Complexity Graph**

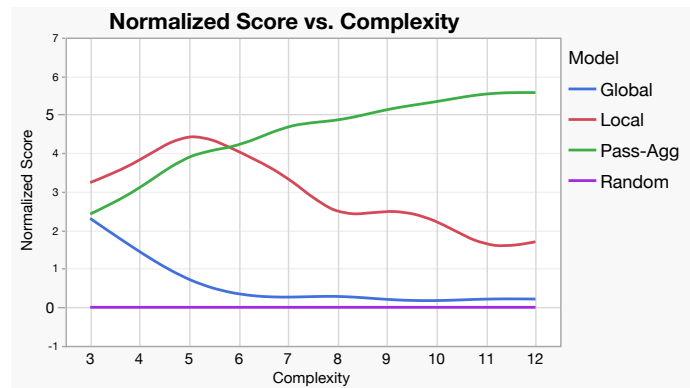

**Figure 8. Normalized Mean Score Vs. Complexity Graph**

As shown in both Figures 7 and 8, *Local* significantly outperforms *Global* across all complexity levels. We see clearly that *Global's* performance begins to decrease from the beginning and then converges to our theoretical zero of *Random*. *Local*, on the other hand, maintains improved performance over *Rule-Based* until complexity level 5, after which it begins to decline in performance until it converges at a level still considerably above *Global*.

Surprisingly, while we expected *Global* to outperform *Local* in the smaller scenarios (e.g., in complexity levels 3 and 4), we found that *Local* outperformed *Global* by a large margin (149.940 points or 475.975% even in the simplest of scenarios). The increased performance of Local over Global, even in the smaller scenarios, may be due to the localized observation always being centered on the agent on-move. This could facilitate learning as this consistent perspective could allow for quicker generalization. Visual replays of these scenarios confirm the better performance of *Local* over *Global* across all scenarios.

We also find that *Local* significantly outperformed *Rule-Based* through complexity level 5, upwards of 262.122%. However, as expected, *Rule-Based* begins outperforming *Local* after a certain level of complexity is reached, where the given training budget is insufficient for an RL model to generalize quickly enough. In our experiment, this crossover point is complexity level 6. Beyond this point, we observe a gradual decline in performance in *Local* as the complexity level increases. Interestingly, whereas *Global* converges to our *Random*, *Local* still significantly improves over *Global*. Based on experiments that involved training in simpler environments, we expect this crossover point to move further to the right as we increase our training budget.

**CONCLUSION AND FUTURE WORK**

Overall, this research presents a compelling case for implementing a localized observation abstraction with some spatial decay component when training models using RL, specifically within environments where spatial relationships may be crucial. Whereas we hypothesized a trade-off space between the global and localized observation approaches, we find that a localized observation with spatial decay consistently outperforms a global observation approach across all levels of complexity examined. The superior performance of using localized observation is particularly striking in the smaller-scale scenarios, as it was anticipated that the global observation approach would be at least as good as the localized approach, if not better. We posit that the efficacy of the localized abstraction approach is likely due to the agent's improved ability to generalize when centered in the observation space, significantly enhancing the learning process and decision-making ability. This approach balanced reducing state-space complexity with the retention of relevant information, thereby better optimizing the agent's performance.

Revisiting Abel's (2020) three desiderata for useful state abstraction (preserving near-optimal behavior, being learnable and computable efficiently, and reducing the time or data needed for effective decision-making), we find our observation abstraction clearly accomplishes all three. Our agents performed better than agents using global observations given a set training budget; our abstraction of the state space proved more efficient than training the agent to reach the same performance threshold using the global observation space; and our abstraction method reduced the time needed for training to reach a desired performance threshold.

The outcomes of this study underscore the potential of localized observation abstractions to become a pivotal component in the application of RL in complex, dynamic environments, such as those encountered in military combat simulations. By demonstrating the limitations of a global observation approach and the advantages of a localized approach, this work paves the way for future investigations into more sophisticated observation abstraction methods to better enable RL scalability.

We will extend the findings of this study by introducing more complex scenarios (e.g., using more types of terrain and units) and increasing the training budget to examine if the same trends hold valid with increasing complexity across these other dimensions. Furthermore, this study informs our current research area of scaling RL to deal with more complex scenarios via hierarchical reinforcement learning (HRL). These results and insights inform how we can better decompose the environment spatially and best explore the nuanced interplay between RL-trained agents and rule-based agents across varying levels of complexity. Such research will refine and generalize the methodologies discussed and contribute significantly to the broader field of AI, offering insights into the scalable training and deployment of RL agents in real-world scenarios.

**REFERENCES**

Abel, D. (2019). A Theory of State Abstraction for Reinforcement Learning. *Proceedings of the AAAI Conference on Artificial Intelligence*, *33*(01), 9876–9877. https://doi.org/10.1609/aaai.v33i01.33019876

Abel, D. (2020). *A Theory of Abstraction in Reinforcement Learning* [Brown University]. https://cs.brown.edu/~mlittman/theses/abel.pdf

Allen, C., Parikh, N., Gottesman, O., & Konidaris, G. (2021). *Learning Markov State Abstractions for Deep Reinforcement Learning*. 35th Conference on Neural Information Processing Systems.

Allen, J. T. (2022). *Enlisting AI in Course of Action Analysis as Applied to Naval Freedom of Navigation Operations* [Naval Postgraduate School]. https://hdl.handle.net/10945/71041

Andersen, P.-A., Goodwin, M., & Granmo, O.-C. (2018). *The Dreaming Variational Autoencoder for Reinforcement Learning Environments* (arXiv:1810.01112). arXiv. http://arxiv.org/abs/1810.01112

Bellman, R. (1954). The Theory of Dynamic Programming. *Bulletin of the American Mathematical Society*, *60*, 503–515. https://apps.dtic.mil/sti/citations/AD0604386

Berner, C., Brockman, G., Chan, B., Cheung, V., Dennison, C., Farhi, D., Fischer, Q., Hashme, S., Hesse, C., Józefowicz, R., Gray, S., Olsson, C., Pachocki, J., Petrov, M., Salimans, T., Schlatter, J., Schneider, J., Sidor, S., Sutskever, I., … Zhang, S. (2019). *Dota 2 with Large Scale Deep Reinforcement Learning*. 66.

Boron, J. A. (2020). *Developing Combat Behavior Through Reinforcement Learning*. Naval Postgraduate School.

Cannon, C. T., & Goericke, S. (2020). *Using Convolution Neural Networks to Develop Robust Combat Behaviors Through Reinforcement Learning*. Naval Postgraduate School.

Darken, C. (2022). *Atlatl*.

Farama Foundation. (2023). *Gymnasium Documentation*. Gymnasium Documentation. https://gymnasium.farama.org

Giunchiglia, F., & Walsh, T. (1992). A theory of abstraction. *Artificial Intelligence*, *57*(2–3), 323–389. https://doi.org/10.1016/0004-3702(92)90021-O

Ho, M. K., Abel, D., Griffiths, T. L., & Littman, M. L. (2019). The value of abstraction. *Current Opinion in Behavioral Sciences*, *29*, 111–116. https://doi.org/10.1016/j.cobeha.2019.05.001

Holcomb, S. D., Porter, W. K., Ault, S. V., Mao, G., & Wang, J. (2018). Overview on DeepMind and Its AlphaGo Zero AI. *Proceedings of the 2018 International Conference on Big Data and Education*, 67–71. https://doi.org/10.1145/3206157.3206174

Jergéus, E., Oinonen, L. K., Carlsson, E., & Johansson, M. (2022). *Towards Learning Abstractions via Reinforcement Learning* (arXiv:2212.13980). arXiv. http://arxiv.org/abs/2212.13980

Li, L., Walsh, T. J., & Littman, M. L. (2016). *Towards a Unified Theory of State Abstraction for MDPs*. 10.

Mnih, V., Kavukcuoglu, K., Silver, D., Rusu, A. A., Veness, J., Bellemare, M. G., Graves, A., Riedmiller, M., Fidjeland, A. K., Ostrovski, G., Petersen, S., Beattie, C., Sadik, A., Antonoglou, I., King, H., Kumaran, D., Wierstra, D., Legg, S., & Hassabis, D. (2015). Human-Level Control Through Deep Reinforcement Learning. *Nature*, *518*(7540), 529–533. https://doi.org/10.1038/nature14236

Puterman, M. L. (1994). *Markov Decision Processes: Discrete Stochastic Dynamic Programming* (1st ed.). Wiley. https://doi.org/10.1002/9780470316887

Raffin, A. (2018, August 21). *Stable Baselines: A Fork of OpenAI Baselines—Reinforcement Learning Made Easy*. https://towardsdatascience.com/stable-baselines-a-fork-of-openai-baselines-reinforcement-learning-made-easy-df87c4b2fc82

Rood, P. R. (2022). *Scaling Reinforcement Learning Through Feudal Muti-Agent Hierarchy*. Naval Postgraduate School.

Sandholm, T. (2015). *Abstraction for Solving Large Incomplete-Information Games*. 5.

Shanahan, M., & Mitchell, M. (2022). *Abstraction for Deep Reinforcement Learning* (arXiv:2202.05839). arXiv. http://arxiv.org/abs/2202.05839

Silver, D., Hubert, T., Schrittwieser, J., Antonoglou, I., Lai, M., Guez, A., Lanctot, M., Sifre, L., Kumaran, D., Graepel, T., Lillicrap, T., Simonyan, K., & Hassabis, D. (2017). *Mastering Chess and Shogi by Self-Play with a General Reinforcement Learning Algorithm* (arXiv:1712.01815). arXiv. http://arxiv.org/abs/1712.01815

Steppa, C., & Holsch, T. L. (2019). HexagDLy—Processing hexagonally sampled data with CNNs in PyTorch. *SoftwareX*, *9*(2352–7110), 193–198. https://doi.org/10.1016/j.softx.2019.02.010

Sutton, R. S., Precup, D., & Singh, S. (1999). Between MDPs and semi-MDPs: A framework for temporal abstraction in reinforcement learning. *Artificial Intelligence*, *112*(1–2), 181–211. https://doi.org/10.1016/S0004-3702(99)00052-1

Van Hasselt, H., Guez, A., & Silver, D. (2016). Deep Reinforcement Learning with Double Q-Learning. *Proceedings of the AAAI Conference on Artificial Intelligence*, *30*(1). https://doi.org/10.1609/aaai.v30i1.10295

Vinyals, O., Babuschkin, I., Czarnecki, W. M., Mathieu, M., Dudzik, A., Chung, J., Choi, D. H., Powell, R., Ewalds, T., Georgiev, P., Oh, J., Horgan, D., Kroiss, M., Danihelka, I., Huang, A., Sifre, L., Cai, T., Agapiou, J. P., Jaderberg, M., … Silver, D. (2019). Grandmaster Level in StarCraft II Using Multi-Agent Reinforcement Learning. *Nature*, *575*(7782), 350–354. https://doi.org/10.1038/s41586-019-1724-z

Whitt, W. (1978). Approximations of Dynamic Programs, I. *Mathematics of Operations Research*, *3*(3), 231–243. https://doi.org/10.1287/moor.3.3.231